\documentclass[11pt]{article}

\usepackage[preprint]{acl}

\usepackage{times}
\usepackage{latexsym}
\usepackage[T1]{fontenc}
\usepackage[utf8]{inputenc}
\usepackage{microtype}
\usepackage{inconsolata}
\usepackage[final]{graphicx}
\setkeys{Gin}{draft=false}

\usepackage{booktabs}
\usepackage{amsfonts}
\usepackage{amsmath}
\usepackage{nicefrac}
\usepackage{enumitem}
\usepackage{xcolor}
\usepackage{pifont}

\title{Interpreting and Steering a Text-to-Speech Language Model\\ with Sparse Autoencoders}

\author{
  Nikita Koriagin\textsuperscript{$\clubsuit$}
  \And
  Georgii Aparin\textsuperscript{\textcolor{red}{\ding{169}}}
  \And
  Nikita Balagansky\textsuperscript{$\clubsuit$}
  \And
  Daniil Gavrilov\textsuperscript{$\clubsuit$}
  \AND
  \normalfont
  \textsuperscript{$\clubsuit$}\,T-Tech
  \qquad
  \textsuperscript{\textcolor{red}{\ding{169}}}\,AI Foundation and Algorithm Lab
}

\begin{document}

\maketitle

\begin{abstract}
Language models increasingly serve as the backbone of text-to-speech (TTS) systems, yet we understand little about the representations they build when text and generated speech tokens share a single residual stream. We train BatchTopK sparse autoencoders on the LM backbone of CosyVoice3 and introduce a modality-aware auto-interp pipeline that labels each feature from where it fires—text-prefix context, 1-second speech clips, or both. The recovered features are interpretable, spanning phonemes, laughter, accent prompts and speaker gender. Steering through the SAE latent space shows these features are causal rather than merely descriptive: targeted interventions raise laughter probability from 0.02 to 0.79, flip perceived speaker gender, and control speech rate while preserving spoken content. SAE features thus serve both as interpretability objects and as  control directions for TTS synthesis.
\end{abstract}

\section{Introduction}

Mechanistic interpretability of large language models has benefited
enormously from sparse autoencoders, which decompose dense,
polysemantic residual-stream activations into sparse, approximately
monosemantic features~\citep{cunningham2023sparse,templeton2024scaling}.
Paired with LLM-based automatic interpretation
(``auto-interp'')~\citep{bills2023language,paulo2024automatically},
SAEs now provide a scalable path toward understanding what individual
components of a text LM compute.

TTS models built on LM backbones present a qualitatively different
setting.
A model such as CosyVoice3~\citep{cosyvoice3} processes a mixed
sequence: an instruction/text prefix followed by discrete 25\,Hz
speech tokens, and generates the latter autoregressively.
The representations it builds may encode syntactic and semantic
information from the text prefix, acoustic and prosodic properties
of the speech being generated, or both—but we currently have no
principled way to identify which features serve which role.

We make three contributions:
\begin{enumerate}[noitemsep]
  \item \textbf{SAE training on a TTS LM.} We train BatchTopK SAEs
        on CosyVoice3 residual-stream activations on
        ${\approx}250$M tokens.
  \item \textbf{Modality-aware auto-interp.} We adapt auto-interp to
        the mixed text--speech sequence: features are labeled from
        prefix-token context, 1-second speech clips, or both,
        depending on where they activate.
  \item \textbf{Layer-wise feature modality analysis.} We categorize
        features by whether they fire on text-position or
        speech-position tokens, exposing how linguistically- and
        acoustically-grounded directions emerge across layers.
\end{enumerate}

\section{Background and Related Work}

\textbf{SAEs for LM interpretability.}
SAEs decompose LM residual streams into sparse, approximately
monosemantic features~\citep{cunningham2023sparse,templeton2024scaling,lieberum2024gemma},
with TopK/BatchTopK variants~\citep{gao2024scaling}.

\textbf{Automatic interpretation.}
\citet{bills2023language} use LLMs to describe activation patterns,
and \citet{paulo2024automatically} score the resulting labels with a
detection-style protocol over activating vs.\ non-activating
examples.
We adapt this protocol from text-only activations to mixed
text/audio evidence.

\textbf{TTS and speech interpretability.}
Probing has examined speech encoders~\citep{pasad2021layer},
SAEs have been applied to discriminative speech and audio foundation
models~\citep{audiosae}, and steering has been applied to control generation in speech models~\citep{xie2025emosteer,aparin2026whisper}. To our knowledge, this is the first SAE analysis of the residual
stream of a generative TTS LM backbone.

\section{Method}

We apply SAEs to the LLM residual stream of CosyVoice3 and route
each feature's strongest activations to text, audio, or mixed
evidence before automatic labeling and held-out evaluation
(pipeline diagram in Appendix~\ref{app:pipeline}).

\subsection{Model and SAE Training}

We use CosyVoice3~\citep{cosyvoice3}, a TTS system whose LM
backbone is Qwen2.5-0.5B (hidden size 896, 28 layers).
The LM receives a text prompt tokenized with a BPE vocabulary and
generates 25\,Hz discrete speech tokens autoregressively.
We train BatchTopK SAEs~\citep{gao2024scaling} at multiple layers
of the LM backbone with dictionary size $d{=}16{,}384$ and $k{=}50$
active features per token, on ${\approx}250$M tokens from the Emilia
dataset~\citep{he2024emilia}.
We use the full layer sweep for modality and reconstruction analyses,
and use layer 20 for the most detailed qualitative examples.
Training uses the standard reconstruction + sparsity objective with
an auxiliary dead-feature loss.

\subsection{Evidence Extraction}

For each SAE feature, we collect its highest-activating token
positions and convert them into modality-specific evidence.
Exact sequence boundaries tell us whether each activation occurred in
the language prefix or in the speech-token segment.
Text activations are represented by a marked token window, while
speech activations are represented by a short audio clip centered on
the active speech token.
This lets the same auto-interp pipeline operate on text-only,
audio-only, and mixed features without forcing all evidence into a
single format.
Implementation details are given in Appendix~\ref{app:protocol}.

\subsection{Feature Modality Tagging}
\label{sec:modality}

CosyVoice3 is trained on randomly-interleaved text and speech
sequences, so absolute token position carries no stable semantic
meaning across samples.
We therefore characterize each feature by the empirical token-type
composition of its strongest activations.
Features whose top examples are mostly speech-token positions are
audio-modal, features whose examples are mostly prefix-token
positions are text-modal, and the remainder are mixed:
\begin{itemize}[noitemsep]
  \item \textbf{Audio-modal}: speech fraction $\geq 0.8$
  \item \textbf{Text-modal}: speech fraction $\leq 0.2$
  \item \textbf{Mixed}: otherwise
\end{itemize}

\subsection{Automatic Labeling}

We label features with a modality-aware prompt to Gemini 3.0
Pro~\citep{gemini2024}.
The evidence shown to the labeler depends on the feature's modality:

\begin{itemize}[noitemsep]
  \item \textbf{Text-modal features} receive text evidence from the
    instruction/transcript prefix: region, token position, activation
    value, and source text context.
  \item \textbf{Audio-modal features} receive only 1-second speech
    clips centered on speech-token activations.
  \item \textbf{Mixed features} receive both text examples and
    speech clips.  We label a mixed feature only when both evidence
    types are present.
\end{itemize}

The prompt asks for a single concise sentence describing the property
consistently associated with the activating evidence.
For text evidence, labels may refer to lexical, punctuation,
language, or prompt-style patterns.
For audio evidence, labels may refer to acoustic, phonetic, or
prosodic properties.
For mixed evidence, the labeler is asked to describe the cross-modal
relation when one is visible. For exact prompts see Appendix \ref{app:prompts}.

\subsection{Detection-Style Evaluation}

We evaluate labels with a detection-style task
(cf.\ \citealt{paulo2024automatically}) adapted to multimodal
evidence.
Given a proposed label, a held-out evaluator scores shuffled positive
and negative evidence items for how well they match the label.
We report AUROC and balanced accuracy over these held-out judgments,
with balanced feature samples for the text, audio, and mixed groups.
The scoring protocol is described in Appendix~\ref{app:protocol}.

\subsection{SAE Feature Steering}
\label{sec:steering}

To test whether interpreted SAE features can causally control TTS generation, we intervene through the SAE latent space rather than directly adding a residual vector.
During generation, speech-token residual states are encoded by the frozen SAE, selected feature activations are shifted by a signed and normalized amount, and the modified latent state is decoded back into the model residual stream. The intervention strength during generation is controlled by a scalar $\alpha$, while input-text and speech-prompt tokens are left unchanged. Implementation details are given in Appendix~\ref{app:steering-method}.

\section{Results}

\begin{figure*}[!t]
  \centering
  \includegraphics[width=\textwidth]{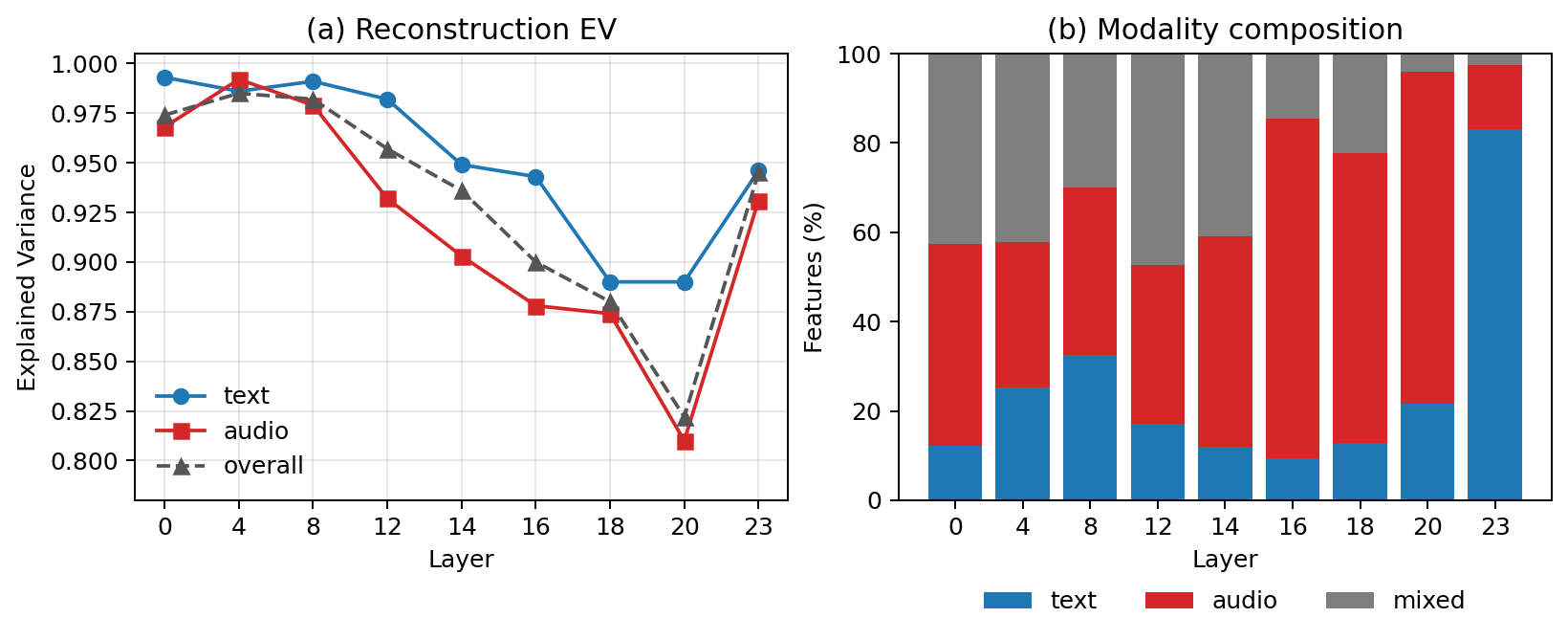}
  \caption{Layer sweep across the Qwen2.5-0.5B backbone.
           \textbf{(a)} SAE explained variance by token type.
           \textbf{(b)} Feature modality composition.
           Middle layers mix text and speech evidence, layers 16--20
           become audio-heavy, and the final hidden state becomes
           mostly text-modal.}
  \label{fig:layer_sweep}
\end{figure*}

\subsection{Layer Sweep}

Figure~\ref{fig:layer_sweep} shows that the SAE dictionary remains a
strong reconstruction model across the TTS backbone while its feature
types shift substantially with depth.
Early and middle layers contain many mixed features, layers 16--20
are dominated by audio-modal features, and the final hidden state
sharply reverts to a mostly text-modal subspace.
This layer-wise movement suggests that the LM backbone does not merely
carry a static text prefix forward: sparse directions become tied to
the generated speech-token stream as acoustic prediction is formed.
Additional layer-sweep details are reported in
Appendix~\ref{app:layer_sweep_details}.

\subsection{Auto-Interp Quality}

\begin{figure}[t!]
  \centering
  \includegraphics[width=\columnwidth]{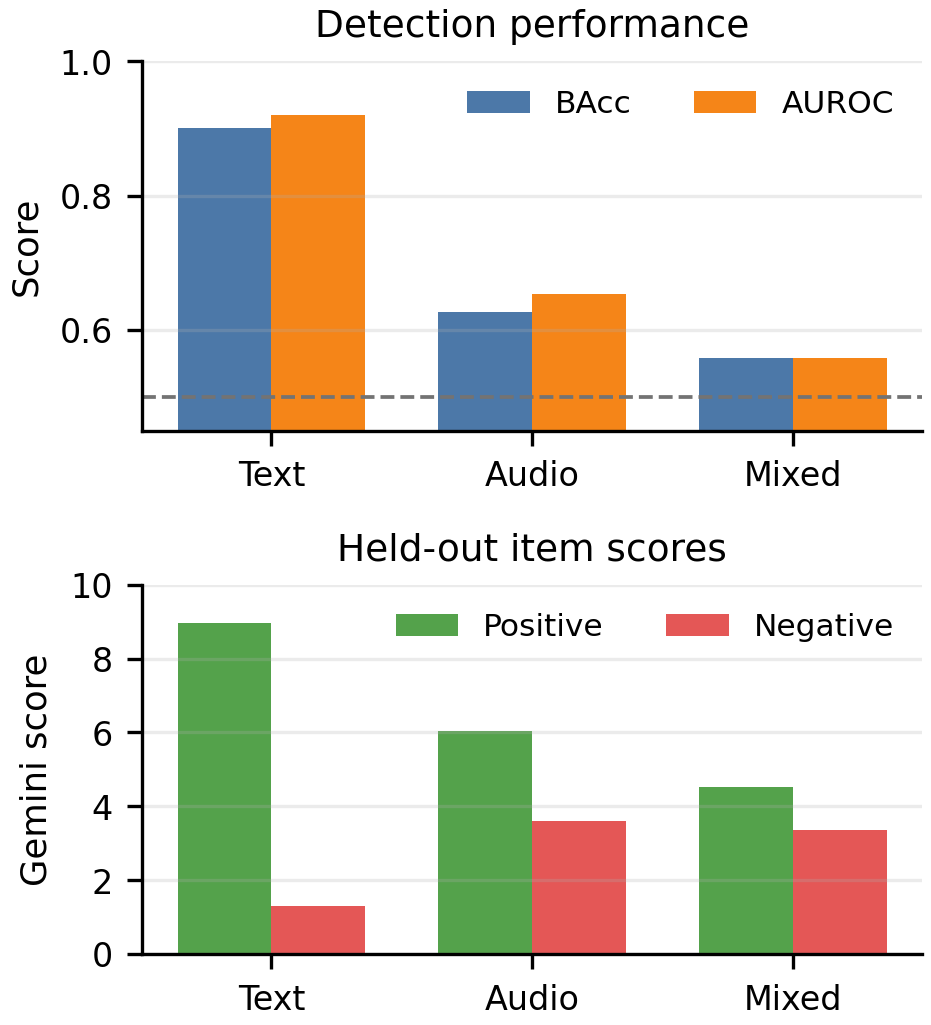}
  \caption{Held-out auto-interp scores for layer-20 features.
           Text labels are easiest to verify; mixed labels are hardest.}
  \label{fig:balanced_eval}
\end{figure}

Figure~\ref{fig:balanced_eval} reports the balanced rank-held-out
auto-interp evaluation for the layer-20 case study.
Text-modal labels are easiest to verify (AUROC 0.921), audio-modal
labels remain above chance (AUROC 0.653), and mixed features
are hardest in aggregate (AUROC 0.558).
The same ordering appears across completed layers
(Appendix~\ref{app:layerwise_scores}).

\begin{table}[t]
  \label{tab:compact_examples}
  \centering
  \small
  \setlength{\tabcolsep}{4pt}
  \begin{tabular}{clp{0.55\columnwidth}c}
    \toprule
    Feat. & Mod. & Auto-interp label & B/A \\
    \midrule
    1376 & Text
      & Word ``British'' in speaker-accent descriptions
      & 1.00/1.00 \\
    233 & Audio
      & The sound of human laughter
      & 0.75/0.75 \\
    5543 & Mixed
      & The phoneme sequence /ohl/ in text and speech
      & 0.92/0.98 \\
    \bottomrule
  \end{tabular}
    \caption{One representative feature per modality. Full examples are
           in Appendix~\ref{app:examples}.}
\end{table}

Qualitative inspection finds text features for individual tokens,
words, years, and voice-prompt attributes. Audio features for phonemes, gender, laughter, stuttering, breaths, and accent cues, and a smaller set of mixed features linking the same word or phoneme-like event across text and speech.
\begin{figure}[t]
\centering
\includegraphics[width=\columnwidth]{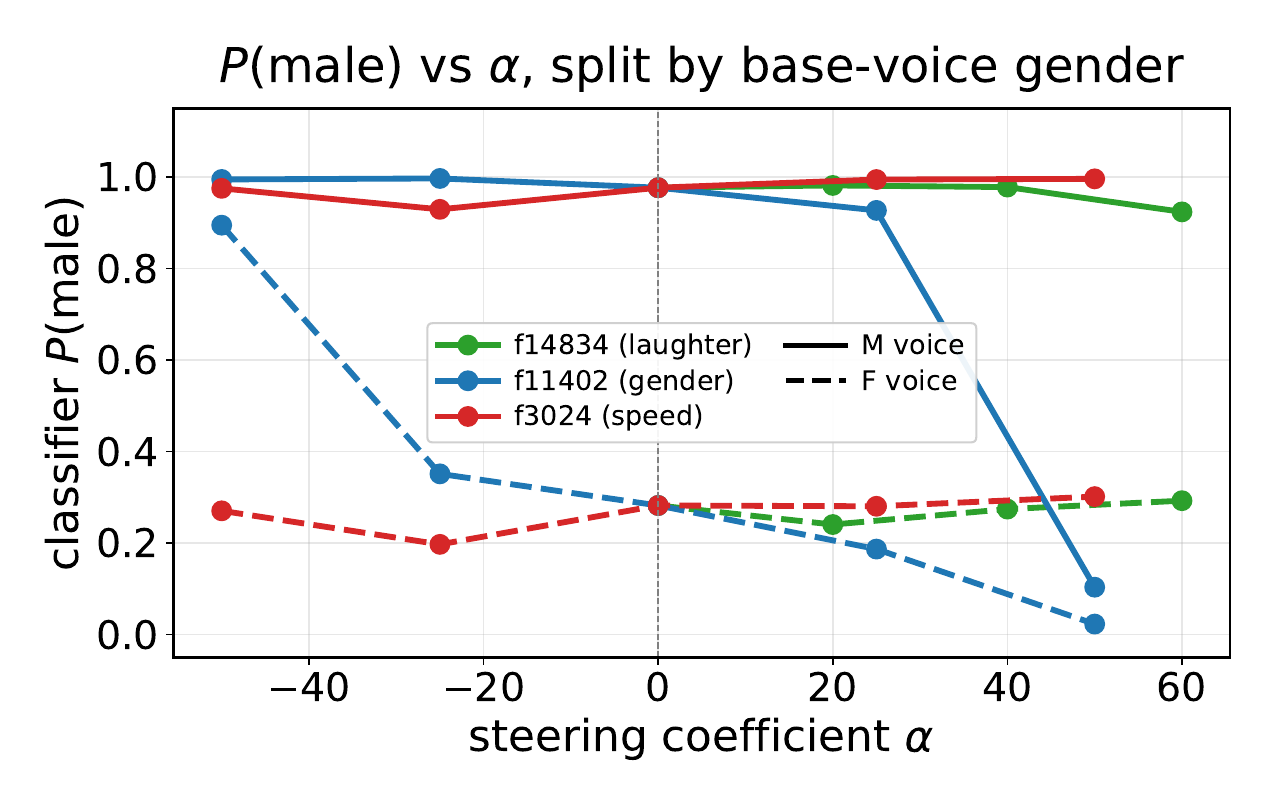}
\caption{
Gender steering split by prompt-speaker gender. Feature 11402 shifts
both male- and female-prompted generations toward the target gender.
}
\label{fig:steering-gender-split}
\end{figure}
\subsection{Probe-Based Feature Selection}
\label{sec:probing}

We use downstream acoustic probes to identify SAE features whose activations are predictive of controllable speech properties. For each candidate feature, we generate a small set of steered samples and score the resulting waveforms with external speech metrics, such as laughter probability, emotion classification and accent classification. The probing protocol and candidate-selection results are reported in
Appendix~\ref{app:concept-probing}.

\subsection{Feature Steering}
\label{sec:steering-results}

\begin{figure}[t]
\centering
\includegraphics[width=0.41\textwidth]{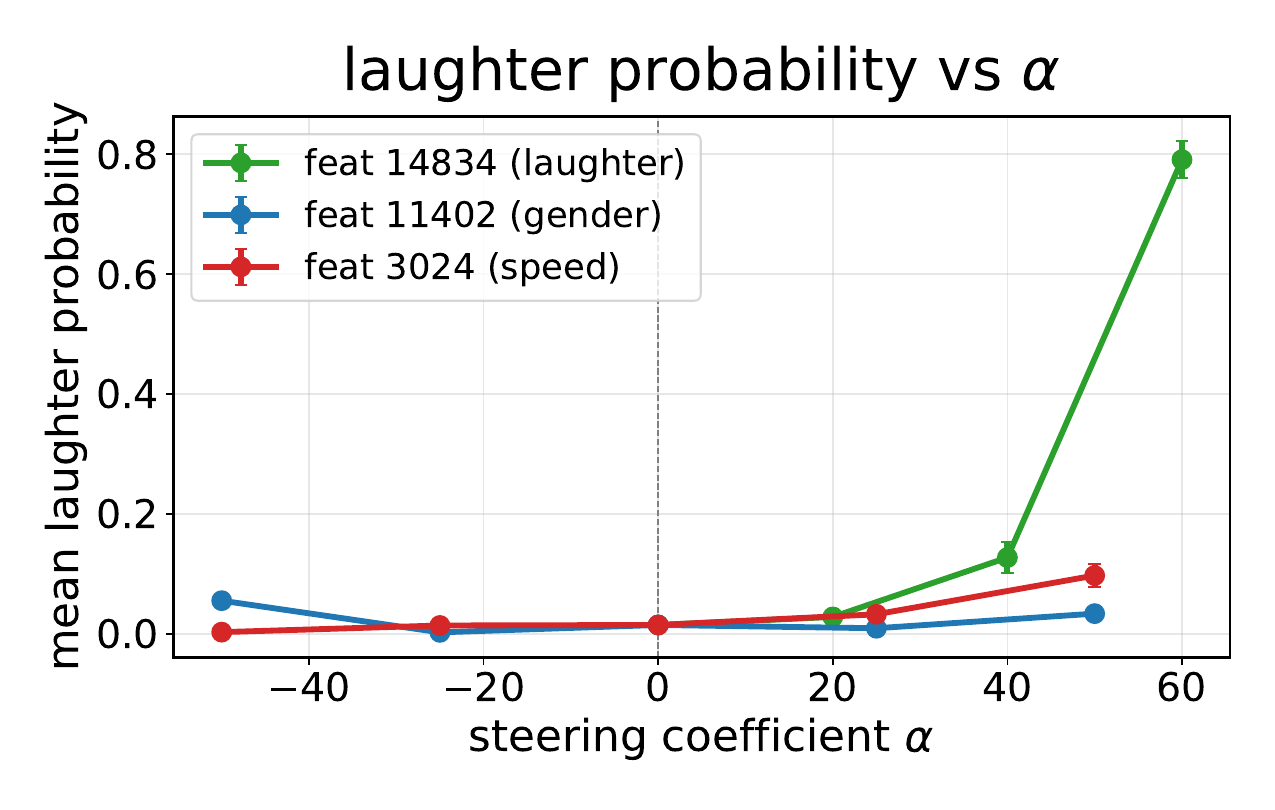}
\hfill
\includegraphics[width=0.41\textwidth]{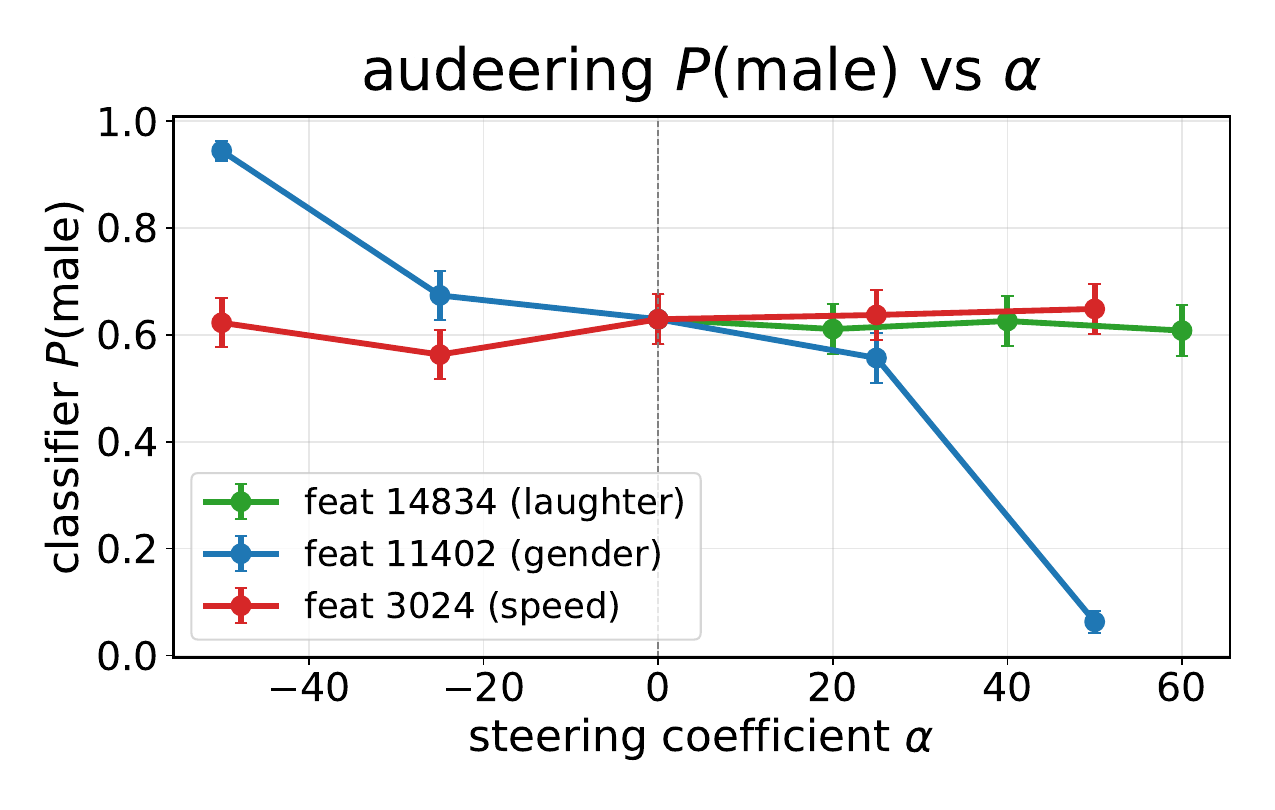}
\hfill
\includegraphics[width=0.41\textwidth]{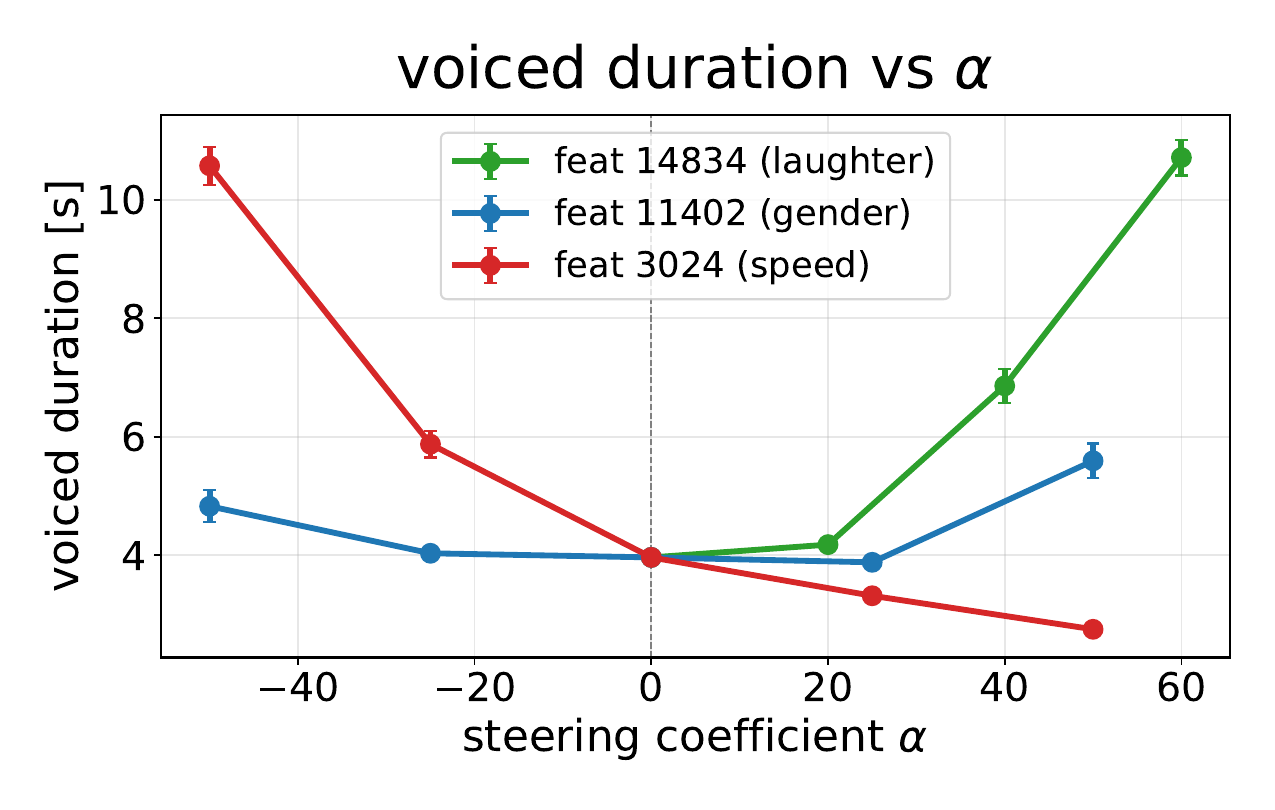}
\caption{
Steering effects for laughter, speaker-gender cues, and speech rate.
Error bars show $\pm 1$ SEM over the $40$-voice $\times$ $10$-text prompts grid.
}
\label{fig:steering}
\end{figure}

We finally test whether interpreted SAE features can be used as causal controls for synthesis. We steer three layer-20 features with labels, shown in Table~\ref{tab:steering-features}. The resulting generations show targeted changes along the corresponding acoustic axes (Figure~\ref{fig:steering}). Feature 14834 increases laughter probability from $0.015$ to $0.791$ at $\alpha=+60$.
Feature 11402 changes speaker-gender cues, moving wav2vec2
$P(\mathrm{male})$ from $0.629$ at baseline to $0.944$ at
$\alpha=-50$ and $0.063$ at $\alpha=+50$. Feature 3024 controls speech
rate, changing voiced duration from $3.96$ s at baseline to $10.57$ s
at $\alpha=-50$ and $2.75$ s at $\alpha=+50$, preserving spoken content.

\begin{table}[t]
\centering
\small
\begin{tabular}{lll}
\hline
Feat. & Auto-interp label & Steering effect \\
\hline
14834 & Laughter-like vocal events & More laughter \\
11402 & Speaker-gender cues & Male/female shift \\
3024 & Speech-rate variation & Slower/faster speech \\
\hline
\end{tabular}
\caption{
Layer-20 SAE features used for steering.
}
\label{tab:steering-features}
\end{table}

Figure~\ref{fig:steering-gender-split} further separates the gender
steering result by the prompt speaker's original gender.

\section{Discussion}

\textbf{What TTS features encode.}
The feature-modality split suggests that CosyVoice3's LM backbone
does more than carry forward the linguistic prefix: by late layers,
many sparse directions are tied to speech-token positions.
Text-modal features capture lexical and prompt-side structure, while
audio-modal features often correspond to phonetic, acoustic, or
prosodic properties.
Mixed features may reflect cross-modal structure involved in mapping
text to speech.

\section{Conclusion}

SAEs trained on a TTS LM recover interpretable text-modal,
audio-modal, and mixed features, and a modality-aware auto-interp
pipeline labels them with descriptions testable by a detection-style
scorer.
Steering experiments further show that some interpreted SAE features are not merely descriptive: intervening on their latent activations can causally control perceptual properties of generated speech. This suggests that SAE features can serve both as interpretability objects
and as practical control directions for TTS synthesis.

\section{Limitations}
\textit{Single model:} results are for CosyVoice3-0.5B and may not
transfer to larger TTS models.
\textit{Circular evaluation:} the labeler and scorer share the same
Gemini model, so systematic hallucinations would inflate scores;
human evaluation and a scorer-model ablation are needed.
\textit{Partial auto-interp sweep:} modality and reconstruction
statistics are reported across layers, while detection-style
auto-interp scores are available for the completed subset of layers.
\textit{Sub-token onset:} speech-token timestamps are at the 25\,Hz
rate and do not localize sub-token (40\,ms) acoustic onsets.
\textit{Negative sampling:} negatives are drawn from other features,
which tests label specificity but not robustness against
representation-neighbor confounds.
\bibliography{references}

\clearpage
\appendix

\section*{Appendix}
\addcontentsline{toc}{section}{Appendix}

\section{Pipeline Diagram}
\label{app:pipeline}

\begin{figure*}[t]
  \centering
  \includegraphics[width=\textwidth]{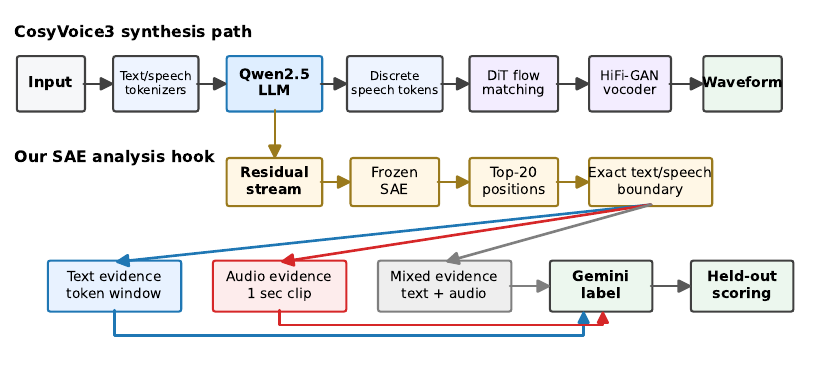}
  \caption{CosyVoice-aware view of our modality-aware SAE
           interpretation pipeline.
           CosyVoice3 synthesizes speech by passing text and prompt
           audio through tokenizers, an LLM backbone, discrete speech
           token generation, DiT flow matching, and a HiFi-GAN vocoder.
           Our analysis attaches SAEs to the LLM residual stream,
           uses exact text/speech boundaries to route top activations
           to text, audio, or mixed evidence, then labels and scores
           the resulting features.}
  \label{fig:method_pipeline}
\end{figure*}

\section{Experimental Protocol Details}
\label{app:protocol}

\paragraph{Activation evidence.}
For each feature, we identify its top-20 activating token positions
across the dataset by encoding residual-stream activations through
the frozen SAE.
The LM is run in teacher-forced mode with sequence layout
\[
  [\mathrm{sos}\ |\ \mathrm{instruct}\ |\ \mathrm{text}\ |
  \mathrm{task}\ |\ \mathrm{speech}],
\]
where the task boundary is a single special token marking the
text-to-speech transition.
For every sample we record the exact instruction, text, and speech
token lengths.
For speech-token activations, positions are mapped to audio timestamps
using the 25\,Hz speech-token rate, and we extract a 1-second window
centered on the peak position from the source audio.
Padded positions are masked before top-activation search, evidence
extraction, and modality statistics.

\paragraph{Modality assignment.}
For each feature we compute the fraction of top-20 activating
positions that fall in the speech segment of their respective
sequences.
Let
$b_i = 1 + \ell^{(i)}_{\mathrm{instruct}} +
        \ell^{(i)}_{\mathrm{text}} + 1$
be the speech-start position for sample $i$.
A top activation at token position $p$ is counted as speech iff
$p \geq b_i$ and $p$ is before the sample's valid sequence length.
Features with speech fraction at least 0.8 are audio-modal; features
with speech fraction at most 0.2 are text-modal; all others are mixed.

\paragraph{Held-out scoring.}
To prevent the scorer from seeing the same examples used to write the
label, we use a rank-held-out split over activation-ranked examples:
the five strongest activations are reserved for labeling and
lower-ranked activations are used for scoring.
For the layer-20 balanced comparison, we evaluate all 668 mixed
features together with matched 668-feature samples of text-modal and
audio-modal features.
Each scoring prompt contains held-out positive evidence from the
target feature and negatives from other features.
The scorer rates each item from 0--10 for how well it matches the
label; we compute AUROC over the resulting positive-vs.-negative
ranking and balanced accuracy after thresholding ratings at 5.

\section{SAE Steering Implementation}
\label{app:steering-method}

We implement steering as an intervention in the latent space of the frozen SAE attached to a transformer residual stream. The hook is registered at the same layer used for the interpreted features and is applied only to speech-token positions. The hook is applied to the speech-prompt segment during prefill and only to the current generated speech-token position during autoregressive decoding. Instruction, text-prefix, and task-token positions are not modified.

For each hooked residual vector $h$, we first compute SAE activations:
\[
z = \sigma(W_{\mathrm{enc}} h + b_{\mathrm{enc}}).
\]
We then perturb selected SAE coordinates by a signed, feature-wise
amount:
\[
z' = z + \alpha \cdot s \odot \bar{Z}.
\]
Here $\alpha$ is the scalar steering strength, $s$ is a sparse sign
vector specifying the polarity of each steered feature, and $\bar{Z}$
contains feature-wise activation scales. Entries outside the selected
feature set are set to zero, so the intervention changes only the
intended SAE coordinates.

Finally, the modified latent vector is decoded back to the model
residual space:
\[
\hat{h}' = W_{\mathrm{dec}} z' + b_{\mathrm{dec}} .
\]
The hook replaces the original residual vector at the selected
speech-token positions with $\hat{h}'$. This makes the intervention
local to the SAE feature subspace while preserving the model's normal
autoregressive generation procedure outside the hooked positions.

\section{Additional Layer-Sweep Details}
\label{app:layer_sweep_details}

\paragraph{Feature modality distribution.}
Figure~\ref{fig:layer_sweep}b shows how the modality composition of
SAE features evolves across the 24-layer Qwen2.5-0.5B backbone.
We identify three regimes.
\textbf{Early and middle layers (0--14)} are dominated by
\emph{mixed} and \emph{audio} features, with text-modal features in
the minority ($12$--$33\%$).  Even at the embedding output (layer 0)
only $12.3\%$ of features are text-modal, while $45.1\%$ are
audio-modal and $42.6\%$ fire on both segments: the residual stream
is already strongly multimodal.  Mixed features are most prominent in
this band, peaking at $47.3\%$ at layer 12, consistent with
cross-modal fusion rather than independent specialization.
\textbf{Late layers (16--20)} are the audio-commitment zone:
audio-modal features dominate ($76.1\%$ at layer 16, $65.0\%$ at
layer 18, $74.3\%$ at layer 20) while mixed features collapse (from
$40.9\%$ at layer 14 to $4.1\%$ at layer 20).  Features attach
decisively to the speech-token segment as the network finalizes the
acoustic prediction.
\textbf{Layer 23} (the final hidden state) reverses sharply: $83.1\%$
text-modal, $14.3\%$ audio-modal, only $2.6\%$ mixed.  Inspecting
per-sample activations, layer 23's sequence is much shorter than the
intermediate layers and its position distribution leans heavily
toward text-prefix tokens, suggesting the final residual stream
re-projects toward a text-vocabulary-aligned subspace before the
output head.

\paragraph{Reconstruction quality per modality.}
Figure~\ref{fig:layer_sweep}a reports the SAE's per-token explained
variance (EV) on a 5{,}000-sample sweep, with positions partitioned
by token type.
EV is computed against the modality-restricted mean, so the text and
audio columns are independent measurements of how well the SAE
reconstructs each subdistribution.
Three trends are visible.
First, reconstruction quality stays high through the early layers
(overall EV $0.97$--$0.99$ at layers 0--8) and then declines through
the body of the network to a minimum of $0.82$ at layer 20.
Second, text-position activations are generally reconstructed at least
as well as audio positions, with the largest text--audio gap at the
audio-commitment layers ($0.065$ at layer 16, $0.080$ at layer 20);
the early layers are the exception, where the gap is small or even
mildly reversed (audio EV $0.992$ vs.\ text $0.986$ at layer 4).
Third, layer 23 breaks the downward trend sharply: its overall EV
rebounds to $0.945$ and the text/audio gap closes to $0.015$.
Audio activations are generally harder to compress with $k=50$ active
features than text activations, consistent with the larger and more
entropic audio-token vocabulary.

\section{Layer-Wise Auto-Interp Scores}
\label{app:layerwise_scores}

\begin{figure*}[t]
  \centering
  \includegraphics[width=0.95\textwidth]{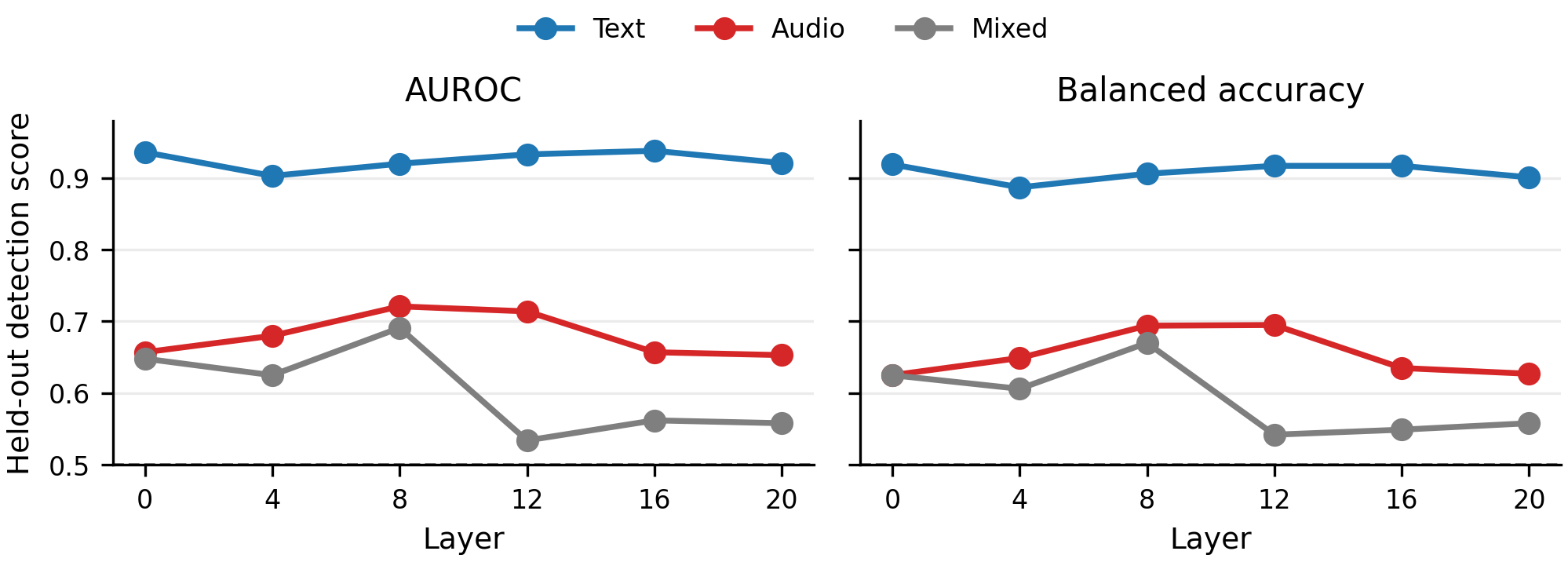}
  \caption{Rank-held-out auto-interp scores across completed layers.
           Text-modal labels remain consistently easiest to verify;
           audio-modal labels are above chance but weaker; mixed labels
           are the most variable.}
  \label{fig:auto_interp_layers}
\end{figure*}

\paragraph{Layer-wise pattern.}
The same ordering appears across the network: text-modal labels are
consistently the easiest to verify (AUROC $0.90$--$0.94$), audio-modal
labels are weaker but reliably above chance (AUROC $0.65$--$0.72$),
and mixed labels are the most variable (AUROC $0.53$--$0.69$).
This supports a conservative interpretation of the qualitative
examples: single-modality features are often well described by one
sentence, while mixed features more often combine several correlated
textual and acoustic properties.

\section{Representative Features}
\label{app:examples}

\begin{table*}[t]
  \caption{Representative layer-20 features with auto-interp labels
           and held-out detection scores.}
  \label{tab:examples}
  \centering
  \small
  \setlength{\tabcolsep}{5pt}
  \begin{tabular}{clp{0.68\textwidth}cc}
    \toprule
    Feat. & Mod. & Auto-interp label & BAcc & AUROC \\
    \midrule
    1376 & Text
      & Word ``British'' in speaker-accent descriptions
      & 1.000 & 1.000 \\
    1443 & Text
      & Morpheme/substring ``ang'' in tokens such as
        ``angry'', ``anger'', ``hanging''
      & 1.000 & 1.000 \\
    1305 & Text
      & Voice prompts describing speaker tone as ``shrill''
      & 1.000 & 1.000 \\
    1330 & Text
      & Four-digit numbers representing years (e.g., 2019, 1936)
      & 1.000 & 1.000 \\
    233 & Audio
      & The sound of human laughter
      & 0.750 & 0.750 \\
    288 & Audio
      & Screaming, shouting, and heavy breathing
      & 0.917 & 1.000 \\
    1225 & Audio
      & Voiceless velar plosive /k/
      & 1.000 & 1.000 \\
    1294 & Audio
      & Phoneme sequence /if/ or /ef/
      & 1.000 & 1.000 \\
    1393 & Audio
      & Spoken /ing/ phoneme sequence
      & 1.000 & 1.000 \\
    164 & Mixed
      & Stutters, false starts, and hesitation markers
      & 0.917 & 1.000 \\
    661 & Mixed
      & The word ``middle'' in text instructions and spoken clips
      & 1.000 & 1.000 \\
    5543 & Mixed
      & The phoneme sequence /ohl/ in text and speech
      & 0.917 & 0.979 \\
    \bottomrule
  \end{tabular}
\end{table*}

\paragraph{Manual observations.}
\textbf{Text features} split into prompt-bound directions (e.g.\ 1376
``British'' and 1305 ``shrill'' in voice-conditioning prompts) and
prompt-independent lexical/sub-lexical patterns (1443 substring
``ang'', 1330 four-digit years), suggesting the dictionary tracks
both the conditioning signal and generic lexical structure.
Most text-modal features are highly local: they activate on
individual BPE tokens, words, punctuation, or short token contexts.
Some correspond to written descriptions of acoustic events or speaker
attributes, such as laughter markers, accent words, and voice-style
adjectives in the instruction.

\textbf{Audio features} span phonemes and short phoneme sequences
(/k/, /if/--/ef/, /ing/), whole-word and non-speech vocal events
(laughter, screams, heavy breathing, stuttering, breaths), and
occasional accent cues.

\textbf{Mixed features} are harder on average and often look
polysemantic, but some are clean: 164 links stutters across transcript
and audio, 661 links the word ``middle'' across text and speech, and
5543 links the same /ohl/-like phoneme sequence in text and speech.
We also observe punctuation-to-pause-like mixed features, but do not
treat these as primary examples unless their held-out scores support
the label.

\section{Concept Probing Experiments}
\label{app:concept-probing}

\paragraph{Setup.}
We run supervised probes for three speech-style concepts: laughter, emotion, and accent.

For each concept and layer, we train a binary logistic-regression probe
\[
\hat{y}=\sigma(\beta^\top \phi(x)+b),
\]
where \(\phi(x)\) is either the raw residual vector \(h_L\in\mathbb{R}^{896}\) or the SAE latent vector \(z_L=\mathrm{SAE}_L(h_L)\in\mathbb{R}^{16{,}384}\). We train probes on the activation from the final speech-prompt token. We use L-BFGS with maximum \(2000\) iterations, apply MaxAbs scaling fit on the training fold only, and report ROC-AUC averaged over stratified 5-fold cross-validation.

\paragraph{Data.}
All three concept probes use the same neutral negative pool: \(500\) Emilia-Yodas neutral-speech clips, filtered to duration \(2\)--\(20\) seconds and DNSMOS at least \(3.0\). We also use a shared text-prefix control: \(500\) LJSpeech transcripts of length \(5\)--\(15\) words, sampled with a fixed seed.

For laughter, positives are \(500\) VocalSound laughter clips. For emotion, positives are \(500\) clips per class from ESD, with one binary probe for each of four emotions: Happy, Sad, Angry, and Surprise. For accent, positives are VCTK 0.92 utterances grouped by speaker accent, with one binary probe per accent. We evaluate eleven accents: English, American, Scottish, Irish, Indian, Canadian, Northern Irish, South African, Australian, Welsh, and New Zealand. The Welsh and New Zealand buckets are single-speaker buckets in VCTK 0.92, so these probes partially conflate accent with speaker identity.

\paragraph{Top-\(k\) SAE feature analysis.}
For each SAE-latent probe, we test whether the concept is distributed across many SAE features or concentrated in a small number of dictionary atoms. We compute the mean coefficient vector \(\bar{\beta}\) across cross-validation folds, select the top-\(k\) SAE coordinates by \(|\bar{\beta}|\), and re-fit the same cross-validated logistic-regression probe using only those coordinates. The main reported values use
\[
k\in\{1,5,10,25,50,100\}.
\]
The \(k=1\) setting is the monosemanticity test: high ROC-AUC with one coordinate means that a single SAE feature carries most of the class-separating signal.

\begin{table*}[t]
\centering
\small
\begin{tabular}{lccccccc}
\hline
Concept / representation & L0 & L4 & L8 & L12 & L16 & L20 & L23 \\
\hline
Laughter, raw residual & 0.885 & 0.980 & 1.000 & 1.000 & 1.000 & 1.000 & 1.000 \\
Laughter, SAE latent & 0.866 & 0.948 & 0.998 & 1.000 & 1.000 & 1.000 & 0.999 \\
Emotion, raw residual mean & 0.966 & 0.998 & 1.000 & 1.000 & 1.000 & 1.000 & 1.000 \\
Accent, raw residual mean & 0.896 & 0.986 & 0.998 & 1.000 & 1.000 & 1.000 & 1.000 \\
\hline
\end{tabular}
\caption{Layer-wise concept decodability. Values are mean ROC-AUC over 5 folds. Emotion is averaged over Happy, Sad, Angry, and Surprise probes; accent is averaged over eleven VCTK accent probes.}
\label{tab:probing-layer-sweep}
\end{table*}

\paragraph{Layer-wise decodability.}
Table~\ref{tab:probing-layer-sweep} shows that all three concepts are linearly decodable early in the network. Raw-residual probes cross \(0.99\) ROC-AUC between layers \(4\) and \(8\), and are essentially saturated from layer \(8\) onward. SAE-latent probes closely track raw-residual probes from layer \(8\), showing that the sparse code preserves the relevant speech-style information while mapping it into dictionary coordinates.

\begin{table*}[t]
\centering
\small
\begin{tabular}{lccccccc}
\hline
Concept & L0 & L4 & L8 & L12 & L16 & L20 & L23 \\
\hline
Laughter & 0.508 & 0.805 & 0.909 & 0.924 & 0.929 & 0.695 & 0.831 \\
Emotion mean & 0.507 & 0.886 & 0.763 & 0.927 & 0.928 & 0.887 & 0.856 \\
Accent mean & 0.510 & 0.762 & 0.925 & 0.923 & 0.835 & 0.789 & 0.836 \\
\hline
\end{tabular}
\caption{Top-1 SAE feature ROC-AUC. This table reports the monosemanticity test: classification using only the single SAE coordinate with the largest mean absolute probe coefficient. Emotion and accent are averaged across their class-specific probes.}
\label{tab:probing-top1}
\end{table*}

\begin{figure*}[t]
\centering
\includegraphics[width=0.72\linewidth]{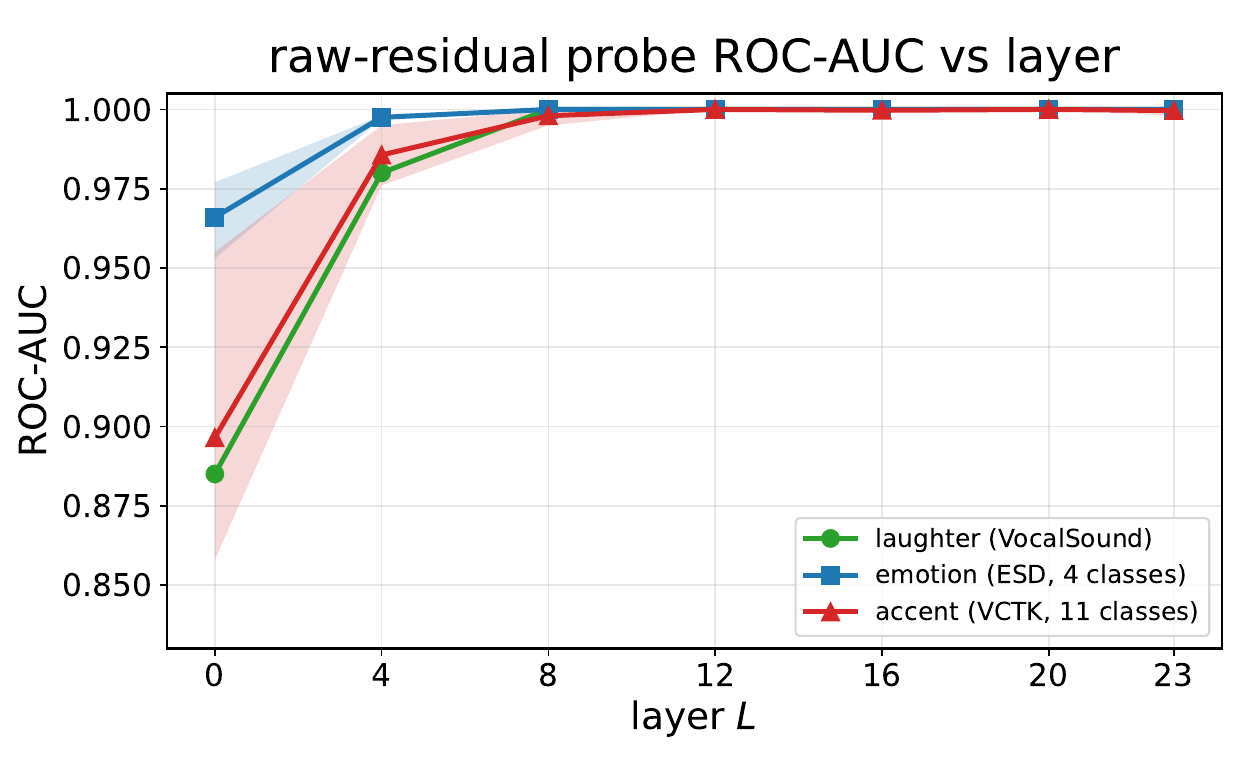}
\caption{Raw-residual probe ROC-AUC as a function of layer for laughter, emotion, and accent. Emotion and accent are shown as mean with min--max bands across their class-specific probes. All three concepts exceed \(0.99\) ROC-AUC by \(L=8\).}
\label{fig:probe-raw}
\end{figure*}

\begin{figure*}[t]
\centering
\includegraphics[width=0.72\linewidth]{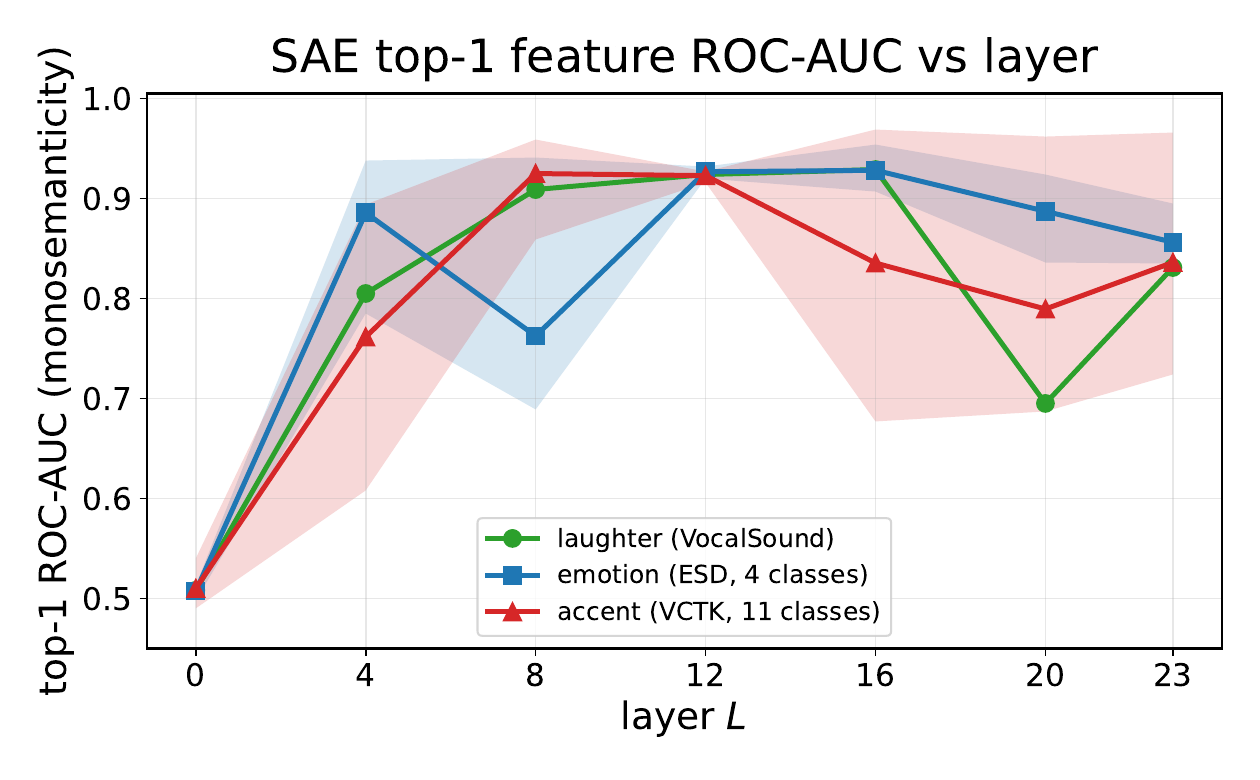}
\caption{Top-1 SAE feature ROC-AUC as a function of layer. This monosemanticity test asks whether a single SAE coordinate is sufficient to separate each concept from neutral speech. Laughter and emotion peak around \(L=12\)--\(16\), while accent is most localized around \(L=8\)--\(12\).}
\label{fig:probe-top1}
\end{figure*}

\paragraph{Layer-wise summary.}
The probing results complement the steering experiments. Probing asks whether a concept is linearly recoverable, and whether it is concentrated in a single SAE coordinate. Figures~\ref{fig:probe-raw} and~\ref{fig:probe-top1} summarize the probing results across layers. The raw-residual probes test concept decodability, while the top-1 SAE probes test whether a single dictionary atom carries the concept. 

\section{Prompts}
\label{app:prompts}

The auto-interp labeling prompt and the detection-style scorer prompt
are reproduced verbatim below.

\subsection{Auto-interp Labeling Prompt}

\begin{quote}\small\ttfamily
You are analyzing internal SAE features in a text-to-speech model.

The feature can activate on text-prefix tokens, speech tokens, or
both. Use all evidence below. If audio clips are present, the
relevant moment is near the middle of the 1-second clip. If text
evidence is present, the active token is marked in each context.
Activation values are normalized to the feature's peak.

Activating text-token evidence: \{text\_evidence\}

Activating audio evidence: \{audio\_evidence\}

Contrast audio clips are unrelated features.

In one concise sentence, describe the property consistently
associated with the activating evidence. Be specific. For text
features, describe the lexical, punctuation, language, prompt-style,
or transcript property. For audio features, describe the acoustic,
phonetic, or prosodic property. For mixed features, describe the
cross-modal relation if one is visible; otherwise say which side
dominates. Do not mention the model or feature.
\end{quote}

\subsection{Detection-Style Scorer Prompt}

\begin{quote}\small\ttfamily
Evaluate whether each of \{n\} evidence items exhibits this SAE
feature:

Feature: ``\{label\}''

Each item may be a text-token context, an audio clip, or both. In
text contexts, the active token is wrapped in [[...]]. In audio
clips, the relevant moment is near the middle of the 1-second clip.

Score each item 0--10:
0--2 : no trace; 3--4 : faint hint; 5--6 : present but not prominent;
7--8 : clearly present; 9--10 : unambiguously prominent.
Use the full range. Do NOT cluster in 2--5.

OUTPUT FORMAT --- your very first line must be the SCORES line,
nothing before it: \mbox{SCORES: [s1, s2, \ldots, sN]}.
Exactly \{n\} integers in item order. After the SCORES line you may
add a brief phrase per item explaining your score.
\end{quote}

\end{document}